\documentclass[10pt,journal]{IEEEtran}
\ifCLASSOPTIONcompsoc
  \usepackage[nocompress]{cite}
\else
  \usepackage{cite}
\fi
\ifCLASSINFOpdf
\usepackage[pdftex]{graphicx}
\graphicspath{{../pdf/}{../jpeg/}}
\DeclareGraphicsExtensions{.pdf,.jpeg,.png}
\else
\usepackage[dvips]{graphicx}
\graphicspath{{images/}}
\DeclareGraphicsExtensions{.eps}
\fi

\usepackage{amsmath}
\usepackage{array}
\usepackage{epsfig}
\usepackage{graphicx}
\usepackage{amsfonts}
\usepackage{float} 
\usepackage{subfigure}
\usepackage{color}
\usepackage{xcolor}
\usepackage{titlesec}
\usepackage{multirow}
\usepackage{rotating}
\usepackage{breakurl}
\usepackage{arydshln}
\usepackage{url}
 
\newcommand{\eg}{\emph{e.g.,}~}
\newcommand{\etal}{\emph{et al.}~}
\newcommand{\ie}{\emph{i.e.,}~}
\newcommand{\todo}[1]{\textcolor{red}{[\textbf{TODO}: #1]}}

\newcommand{\bs}{\boldsymbol}
\usepackage[utf8]{inputenc}
\usepackage{authblk}
\usepackage{comment}
\usepackage{booktabs}
\usepackage{caption}
\begin{document}
\bstctlcite{BSTcontrol}
\title{DGFont++: Robust Deformable Generative Networks for Unsupervised Font Generation}

\author{Xinyuan~Chen,~\IEEEmembership{}
	Yangchen~Xie,~\IEEEmembership{}
	Li~Sun, ~\IEEEmembership{}
	Yue~Lu,~\IEEEmembership{}
	\thanks{X. Chen is with Shanghai Artificial Intelligence Laboratory, Shanghai, China (e-mail:xychen9191@gmail.com).}
	\thanks{Y. Xie, L. Sun, and Y. Lu are with Shanghai Key Laboratory of Multidimensional Information Processing, East China Normal University, Shanghai, 200241, China (e-mail:ycxie0702@126.com;sunli@ee.ecnu.edu.cn;ylu@cs.ecnu.edu.cn)
	}
}
\markboth{}%
{Shell \MakeLowercase{\textit{et al.}}: Bare Demo of IEEEtran.cls for IEEE Journals}
\IEEEtitleabstractindextext{
\begin{abstract}
Automatic font generation without human experts is a practical and significant problem, especially for some languages that consist of a large number of characters. Existing methods for font generation are often in supervised learning. They require a large number of paired data, which are labor-intensive and expensive to collect. In contrast, common unsupervised image-to-image translation methods are not applicable to font generation, as they often define style as the set of textures and colors. In this work, we propose a robust deformable generative network for unsupervised font generation (abbreviated as DGFont++). We introduce a feature deformation skip connection (FDSC) to learn  local patterns and geometric transformations between fonts. The FDSC predicts pairs of displacement maps  and  employs  the  predicted  maps  to apply deformable convolution to the low-level content feature maps. The outputs of FDSC are fed into a mixer to generate final results. Moreover, we introduce contrastive self-supervised learning to learn a robust style representation for fonts by understanding the similarity and dissimilarities of fonts. To distinguish different styles, we train our model with a multi-task discriminator, which ensures that each style can be  discriminated  independently. 
In addition to adversarial loss, another two reconstruction losses are adopted to constrain the domain-invariant characteristics between generated images and content images. Taking advantage of FDSC and the adopted loss functions, our model is able to maintain spatial information and generates high-quality character images in an unsupervised manner. Experiments demonstrate that our model is able to generate character images of higher quality than state-of-the-art methods.
\end{abstract}
}
\maketitle
\IEEEdisplaynontitleabstractindextext
\IEEEpeerreviewmaketitle
\section{Introduction}
 
\begin{figure*}[ht]
\begin{center}
   \includegraphics[width=\linewidth]{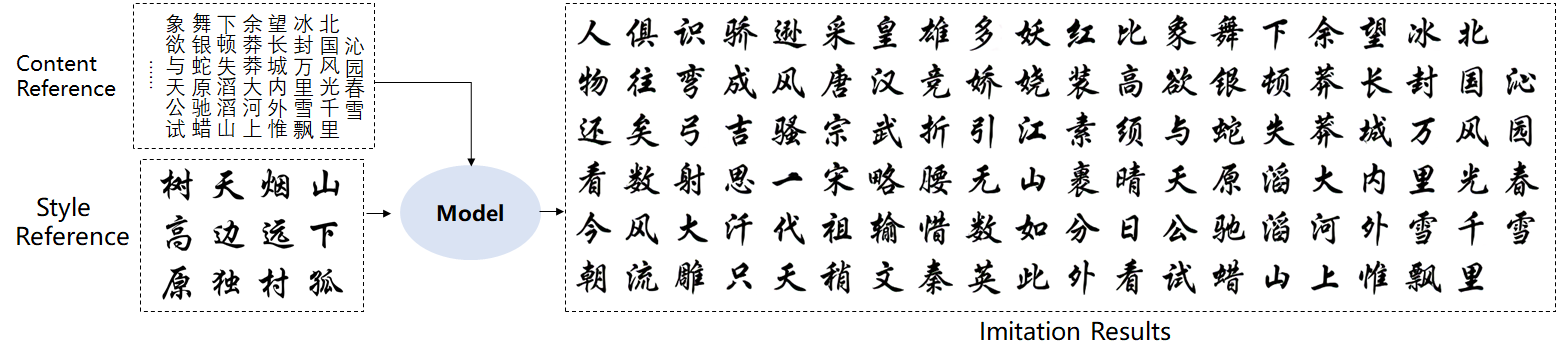}
\end{center}
   \caption{\textbf{Unsupervised font generation results.} Given content images and a few style reference images, our model aims to generate imitations. In this example, the reference images are of a calligraphic font, and the imitation result is generated from our model.} \label{fig:1}
\end{figure*}

Every day, people consume a massive amount of text for information transfer and storage. As the representation of texts, the font is closely related to our daily life. Font generation is critical in many applications, \eg~font library creation, personalized handwriting, historical handwriting imitation, and data augmentation for optical character recognition and handwriting identification. Traditional font library creating methods heavily rely on expert designers by drawing each glyph individually, which is especially expensive and labor-intensive for logographic languages such as Chinese (more than 60,000 characters), Japanese (more than 50,000 characters), and Korean (11,172 characters). 

Recently, the development of convolutional neural networks enables automatic font generation without human experts. There have been some attempts to explore font generation and achieve promising results. \cite{DBLP:journals/corr/UpchurchSB16,DBLP:conf/cvpr/AzadiFKWSD18,DBLP:conf/cvpr/FogelACML20} utilize deep neural networks to generate entire sets of letters for certain alphabet languages. Two notable projects, ``Rewrite" \cite{Rewrite} and ``zi2zi" \cite{zi-2-zi}, generate logographic language characters by learning a mapping from one style to another with thousands of paired characters. After that, EMD \cite{EMD} and SA-VAE \cite{Sun2018LearningTW} design neural networks to separate the content and style representation, which can extend to generate the character of new styles or contents. However, these methods are in supervised learning and required a large amount of paired training samples.

Some other methods exploit auxiliary annotations (\eg strokes, radicals) to facilitate high-quality font generation. For example, \cite{SCFont} utilizes labels for each stroke to generate glyphs by writing trajectories synthesis. \cite{RD-GAN} employ the radical decomposition (\eg radicals or sub-glyphs) of characters to achieve font generation for certain logographic language. DM-Font \cite{Few-shot} and its improved version LF-Font \cite{DBLP:journals/corr/abs-2009-11042} propose disentanglement strategies to disentangle complex glyph structures, which help capture local details in rich text design. MXFont \cite{Park_2021_ICCV} extracts localized features for  few-shot font generation by exploiting sub-glyph and components of characters. However, these methods rely on prior knowledge and can only apply to specific writing systems. Some labels such as the stroke skeleton can be estimated by algorithms, but the estimation error would decrease the generated quality. Also, these methods still require thousands of paired data and annotated labels for training. Recently, there are some attempts \cite{DBLP:conf/aaai/0006W20,DBLP:conf/wacv/ChangZPM18} for unsupervised font generation. \cite{DBLP:conf/wacv/ChangZPM18} introduces a novel module that transfers the features across sequential DenseNet blocks \cite{DBLP:conf/cvpr/HuangLMW17}. \cite{DBLP:conf/aaai/0006W20} proposes a fast skeleton extraction method to obtain the skeleton of characters, and then utilize the extracted skeleton to facilitate font generation.

In the field of image-to-image translation, a series of unsupervised generative models have been proposed by combining adversarial training \cite{Kim_2017_ICML,Yi_2017_ICCV} with carefully designed constraints \cite{Zhu2017UnpairedIT,DBLP:journals/corr/TaigmanPW16,benaim2017one,baek2021rethinking}. However, classical unsupervised image-to-image translation methods cannot be directly applied to unsupervised font generation tasks. In image-to-image translation, the style of images is usually defined as the set of textures and colors. In contrast, different fonts have 
their own local patterns such as stroke thickness, tips of brushes, and joined-up writing patterns. Existing methods for image-to-image translation usually extract the style feature of the target class images and employ adaptive instance normalization (AdaIN) \cite{DBLP:conf/iccv/HuangB17,DBLP:conf/iccv/0001HMKALK19} to combine the content and the style features.
The AdaIN-based methods transfer style by aligning feature statics, which tends to transform texture and color, which is not suitable to transform local style patterns (\eg geometric deformation) for the font. 

 Compelled  by  the  above  observations, we propose a robust deformable generative model for unsupervised font generation (DGFont++). 
The proposed method is designed to deform and transform the character of one font to another by leveraging the provided images of the target font. Figure \ref{fig:1} shows our scenario for font generation. Given a few style reference images (\eg~calligraphy from an artist), our model is able to generate imitations by transforming content reference character images.
Our proposed DGFont++ separates style and content respectively and then mixes two representations to generate target characters. We introduce a feature deformation skip connection (FDSC) to predict pairs of displacement maps and employ the predicted maps to apply deformable convolution to the low-level feature maps from the content encoder. The outputs of FDSC are then fed into a mixer to generate the final results.  
Also, to learn a robust representation for fonts, we use contrastive learning for our style encoder to understand the similarity and dissimilarities of fonts. To this end, we define several data augmentation operations to construct positive counterparts for fonts. The model is then imposed to learn a feature space where characters and their positive counterparts are at a similar point. To distinguish different styles, we train our model with a multi-task discriminator, which ensures that each style can be discriminated independently. In addition to adversarial loss, another two reconstruction losses are adopted to constrain the domain-invariant characteristics between generated images and content images.

The feature deformation skip connection (FDSC) module is used to transform the low-level feature of content images, which preserves the pattern of character (\eg~strokes and radicals). Different from the image-to-image translation problem that defines style as a set of textures and colors, the style of font is basically defined as geometric transformation, stroke thickness, tips of brushes, and joined-up writing pattern. Two fonts with the same content, usually have correspondence for each stroke. Taking advantage of the spatial relationship of fonts, the feature deformation skip connection (FDSC) is used to conduct spatial deformation, which effectively ensures the generated image has complete structures. Moreover, to further improve the quality of generated images, we integrate local spatial attention into an FDSC module, which predicts the local relationship between the encoder and mixer features. The local spatial attention model predicts similarity scores for features of each position regarding its neighboring positions. 

 This work is an extensive version of our conference paper \cite{xie2021dg}.
Compared to the previous version, this paper includes the following additional contributions. (1) A more robust style feature extraction approach is proposed. We introduce data augmentation operations for fonts to construct positive counterparts of characters and introduce contrastive loss to help the model to learn better representation. (2) We integrate local spatial attention into an FDSC  module, \ie~FDSC-attn. Experiments prove the superiority of FDSC-attn compared to Vanilla FDSC. (3) Additional experiments are conducted to ablate and analyze the function of our DGFont++. (4) We conduct more comprehensive experiments of different image sizes and comparisons to state-of-the-art methods.
Extensive experiments demonstrate that our model outperforms state-of-the-art font generation methods. Besides, results show that our model is able to extend to generate unseen style character.
\section{Related Work}
\subsection{Font Generation}
Font generation aims to automatically generate characters in a specific font and create a font library. Recent studies have employed image translation methods for font generation. ``Zi2zi" \cite{zi-2-zi} and ``Rewrite" \cite{Rewrite} implement font generation on the basis of GAN \cite{DBLP:conf/nips/GoodfellowPMXWOCB14} with thousands of character pairs for strong supervision. After that, a series of models are proposed to improve the generated quality based on zi2zi \cite{zi-2-zi}. PEGAN \cite{Sun2018PyramidEG} sets up a multi-scale image pyramid to pass information through refinement connections. HAN \cite{Hierarchical} improves zi2zi by designing a hierarchical loss and skip connection. AEGG \cite{Lyu2017AutoEncoderGG} adds an additional network to refine the training process. DC-Font \cite{Jiang2017DCFontAE} introduces a style classifier to get a better style representation.
However, all the above methods are in supervised learning and require a large number of paired data. In unsupervised font generation, \cite{DBLP:conf/wacv/ChangZPM18,DBLP:conf/aaai/0006W20}  achieve unsupervised font generation by learning a mapping between two fonts directly. However, they ignore the geometric deformation of the font, and their results are not satisfying.

Lots of methods employ auxiliary annotations (\eg stroke and radical decomposition) to further improve the generation quality. SA-VAE \cite{Sun2018LearningTW} disentangles the style and content as two irrelevant domains with encoding Chinese characters into high-frequency character structure configurations and radicals. CalliGAN \cite{CalliGAN} further decomposes characters into components and offers low-level structure information including the order of strokes to guide the generation process. RD-GAN \cite{RD-GAN} proposes a radical extraction module to extract rough radicals which can improve the performance of the discriminator and achieves the few-shot Chinese font generation.  DM-Font \cite{Few-shot} and its improved version LF-Font \cite{DBLP:journals/corr/abs-2009-11042} propose disentanglement strategies to disentangle complex glyph structures, which help capture local details in rich text design. MXFont \cite{Park_2021_ICCV} extracts localized features for  few-shot font generation by exploiting sub-glyph and components of characters.
Some other attempts have been made in Chinese character generation by adopting skeleton/stroke extraction algorithm \cite{DBLP:conf/aaai/0006W20,SCFont}. However, they need extra annotations or algorithms to guide font generation; while the estimation error would decrease the generation performance. In this work, our model, DGFont++, aims to generate high-quality character images in an unsupervised way without other annotations.
\subsection{Image-to-Image Translation}
The purpose of image-to-image translation is to learn a mapping from an image in the source domain to the target domain. Image-to-image translation has been applied in many fields such as artistic style transfer \cite{Johnson2016PerceptualLF,Zhang2018MultistyleGN}, semantic segmentation \cite{Shukla2019ExtremelyWS,Musto2020SemanticallyAI}, image animation \cite{9229197,yang2017pairwise,Chen_2020_CVPR}, object transfiguration \cite{chen2018attention}, and video frames generation \cite{chan2019everybody,9107481,dong2021dual} \etal
Pix2pix \cite{Isola2017ImagetoImageTW} is the first model proposed for image-to-image translation based on conditional GAN \cite{Mirza2014ConditionalGA}. 
To achieve unsupervised image-to-image translation, a lot of works \cite{Liu2016CoupledGA,Zhu2017UnpairedIT,Bousmalis2017UnsupervisedPD,Shrivastava2017LearningFS} have been proposed, where Cycle-GAN \cite{Zhu2017UnpairedIT} introduces a cycle consistency between source and target domain to discover the relationship of samples between two domain. However, the above-mentioned methods can only translate from one domain to another specific domain. To tackle this problem, recent works \cite{DBLP:journals/tip/ChenXYST19,DBLP:conf/iccv/0001HMKALK19,baek2021rethinking,DBLP:conf/cvpr/BhattacharjeeKV20} are proposed to simultaneously generate multiple style outputs given the same input. Gated-GAN \cite{DBLP:journals/tip/ChenXYST19} proposes a gated transformer to transfer multiple styles in a single model. FUNIT \cite{DBLP:conf/iccv/0001HMKALK19} encodes content image and class image respectively, and combines them with AdaIN \cite{DBLP:conf/iccv/HuangB17}. TUNIT \cite{baek2021rethinking} further introduces a guiding network as an unsupervised domain classifier to automatically produce a domain label of a given image. DUNIT \cite{DBLP:conf/cvpr/BhattacharjeeKV20} extracts separate representations for the global image and for the instances to preserve the detailed content of object instances. To learn the mapping across geometry variations, \cite{ganimorph} introduces a discriminator with dilated convolutions as well as a multi-scale perceptual loss that can represent errors in the underlying shape of objects. \cite{transgaga} disentangles image space into a Cartesian product of the appearance and the geometry latent spaces.

Our task is related to unsupervised image translation \cite{DBLP:conf/eccv/HuangLBK18, DBLP:conf/iccv/0001HMKALK19, baek2021rethinking}, which shares the same aim to translate images from one domain to another based on reference images of the target domain. However, these classical unsupervised image-to-image translation methods usually focus on images of wild animals or style transfer whose style is defined as poses or a set of textures and colors. There exist some works for improving shape deformation in the unsupervised image-to-image translation, such as GAN-imorph \cite{ganimorph}, but they still cannot generate high-quality font images. In contrast, characteristics of font lie in local patterns such as stroke thickness, tips of brushes, geometric deformation, and joined-up writing patterns. The unique characteristics of the font motivate the design of robust deformable networks for font generation. 
\subsection{Deformable Convolution and Attention Mechanism}
CNNs have inherent limitations in modeling geometric transformations due to the fixed kernel configuration. To enhance the transformation modeling capability of CNNs, \cite{Deformablev1} proposes the deformable convolutional layer. It augments the spatial sampling locations in the modules with additional offsets. The deformable convolution has been applied to address several high-level vision tasks, such as object detection \cite{Bertasius2018ObjectDI,Deformablev1,Deformablev2} video object detection 
\cite{DBLP:conf/icassp/ChenLF0Y20} sampling, semantic segmentation \cite{Deformablev2}, and human pose estimation \cite{DBLP:conf/eccv/SunXWLW18}. Recently, some methods attempt to apply deformable convolution in image generation tasks. TDAN \cite{DBLP:conf/cvpr/TianZ0X20} addresses video super-resolution tasks by using deformable convolution to align two continuous frames and output a high-resolution frame. \cite{DBLP:conf/eccv/YinSL20, DBLP:conf/cvpr/YinSL21} synthesized novel  view images by deformable convolution given the view condition vectors. In our proposed DGFont, offsets are estimated by a learned latent style code.

 Furthermore, attention modules and deformable convolution can be viewed from a unified perspective \cite{DBLP:conf/iccv/ZhuCZLD19}. The proper combination of deformable convolution and attention module achieves higher accuracy than using one of them in the tasks of object detection and semantic segmentation. \cite{DBLP:conf/cvpr/YinSL21} proposes Soft and Hard Conditional Deformation Modules which employ deformable convolution and attention in a special skip connection way in supervised image synthesis tasks. They show that attention is more effective when applied to lower-resolution features, which is also demonstrated by \cite{DBLP:conf/icml/ZhangGMO19, DBLP:journals/corr/abs-2103-16748}. Our proposed model shows that the combination of deformable convolution and spatial attention works well in unsupervised font generation tasks.
\begin{figure*}
\begin{center}
\includegraphics[width=0.9\textwidth]{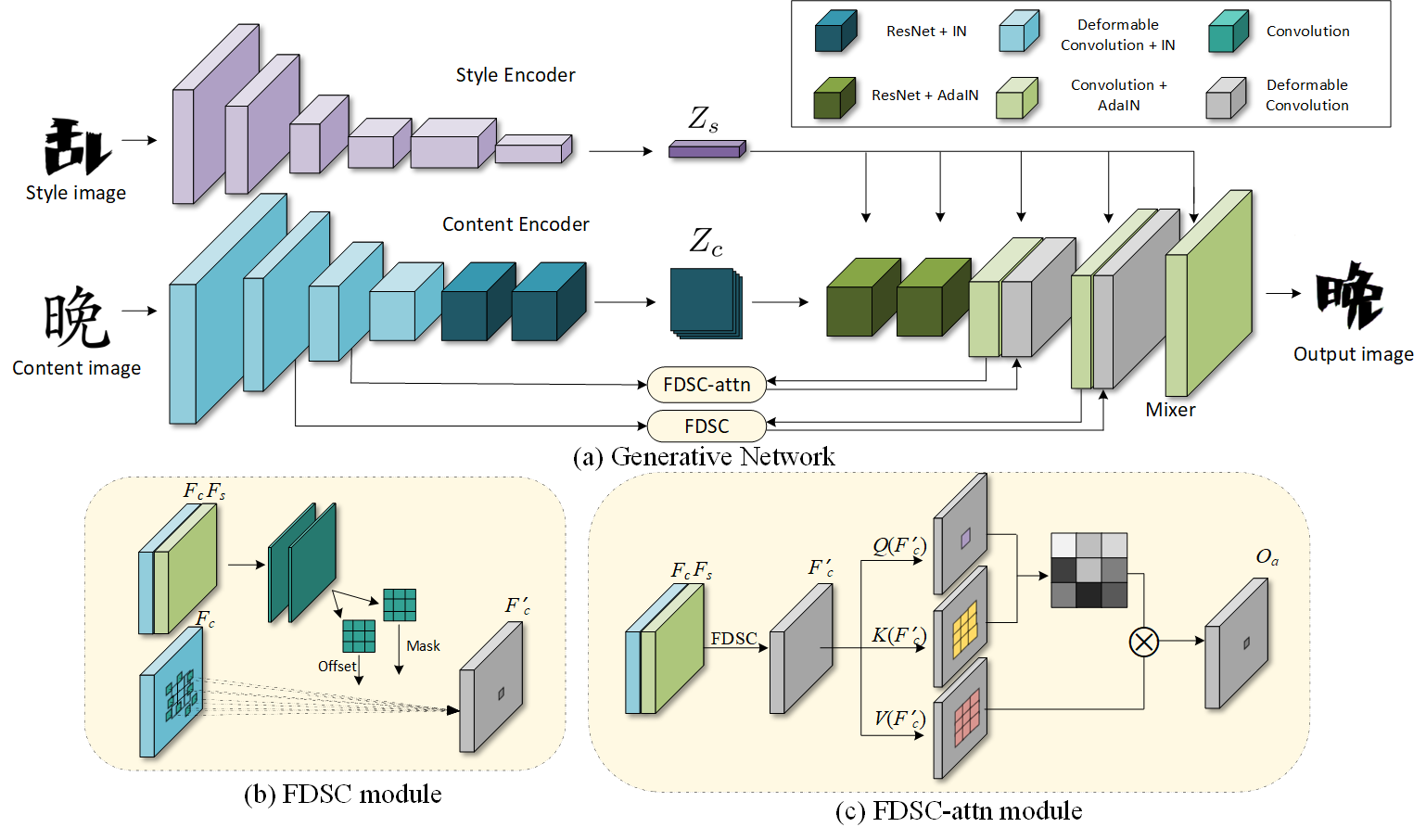}
\end{center}
   \caption{\textbf{Overview of the proposed method}. a) Overview of our generative network. The style/content encoder maps style/content image to style/content representation $Z_s$/$Z_c$. FDSC and FDSC-attn apply transformation convolution to the low-level feature from the content encoder and inject the results into the mixer. The mixer generates the output image. b) A detailed illustration of the FDSC module. c) A detailed illustration of the FDSC-attn module.}
\label{fig:2}
\end{figure*}
 \subsection{Unsupervised Representation Learning}
In this study, we utilize unsupervised learning to learn a robust style feature representation. Unsupervised representation learning aims to extract meaningful representations for downstream tasks without human supervision. Recently, a family of contrastive learning methods based on maximizing mutual information has been proposed \cite {DBLP:conf/icml/ChenK0H20, DBLP:conf/nips/ChenKSNH20, DBLP:conf/cvpr/He0WXG20, DBLP:journals/corr/abs-2003-04297, DBLP:conf/icml/Henaff20, DBLP:conf/iclr/HjelmFLGBTB19,DBLP:conf/eccv/ParkEZZ20}. These methods make use of noise contrastive estimation \cite{DBLP:journals/jmlr/GutmannH10}, learning an embedding where associated samples are brought together, in contrast to other samples in the dataset. The design choices of the contrastive loss, such as the number of negatives and how to sample them, and data augmentation all play a critical role and need to be carefully studied.
By maintaining the dictionary as a queue of negative data samples and employing the data augmentation images as positive samples, Moco \cite{DBLP:conf/cvpr/He0WXG20} achieves outstanding performance in various downstream tasks under a reasonable mini-batch size on the ImageNet dataset. Due to the gap between natural images and font images, the data augmentation used in Moco cannot be directly applied to our framework. We adopt contrastive learning in our framework and design several data augmentations for font images.
\section{Methods}
\subsection{Overview}
Given a content image $I_c$ and a style image $I_s$, our model aims to generate the character of the content image with the font of the style image. As illustrated in Figure \ref{fig:2}, the proposed generative network consists of a style encoder, a content encoder, a mixer, and two feature deformation skip connection (FDSC) modules. The architecture of the style encoder and discriminator is simplified in Figure \ref{fig:2}. The detailed architecture is in Section \ref{sec:arc}. \textbf{The style encoder} is designed to learn the style representation from input images. The style encoder takes a style image as the input and maps it to a style latent vector $Z_{s}$. 
\textbf{The content encoder} is introduced to extract the structure feature of the content images. The content encoder maps the content image into a spatial feature map $Z_{c}$. The content encoder module is made of three deformable convolution layers followed by two residual blocks. The introduced deformable convolution layer enables the content encoder to produce style-invariant features for images with the same content. 
\textbf{The mixer} aims to output characters by mixing the content feature representations $Z_c$ and style feature representations $Z_s$. AdaIN \cite{DBLP:conf/iccv/HuangB17} is adapted to inject the style feature into the mixer. Besides, the \textbf{feature deformation skip connection (FDSC)} modules transfer the deformed low-level feature from the content encoder to the mixer. Details are described in Sec \ref{sec:deform_module}.
\begin{figure}[t]
\begin{center}
   \includegraphics[width=0.9\linewidth]{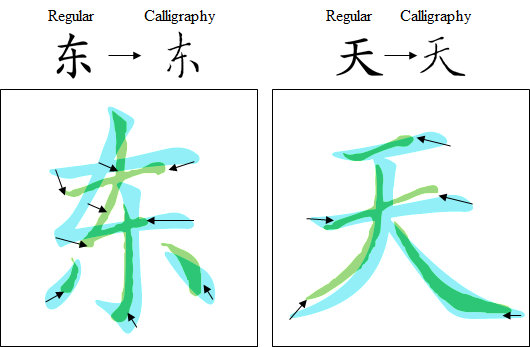}
\end{center}
   \caption{\textbf{The geometric deformation of two fonts for a character}. There is a correspondence for each stroke between two fonts for the same character.}
\label{fig:3}
\end{figure}
\subsection{Feature Deformation Skip Connection}
\label{sec:deform_module}
As illustrated in Figure \ref{fig:3}, there lies a geometric deformation of two fonts for a character and exists a correspondence for each stroke. Compelling this observation, we propose a feature deformation skip connection (FDSC) module to apply geometric deformation convolution to the content image in the feature space and directly transfer the deformation low-level feature to the mixer. Specifically, the module predicts offsets based on the guidance code to instruct the deformable convolution layer to perform a geometric transformation on the low-level feature. As demonstrated in Figure \ref{fig:2}, the input of the FDSC module is a concatenation of two feature maps: a feature map $F_c$ extracted from the content image and a style guidance map $F_s$. $F_s$ is extracted from the mixer after injecting the style code $F_s$. The module estimates sampling parameters after applying convolution to the concatenation of $F_s$ and $F_c$:
\begin{equation}
\Theta = f_{\theta}(F_s, F_c).
\end{equation}
Here, $f_{\theta}$ refers to a deformable convolution layer, and $\Theta$ = \{$\Delta p_{k}$, $\Delta m_{k}$ $|\  k$ = 1, $\cdots,|\mathcal{R}|$\} refers to the offsets and mask of the convolution kernel, where $\mathcal{R}$ = \{(-1, -1), (-1, 0), $\cdots$, (0, 1), (1, 1)\} indicates a regular grid of a 3$\times$3 kernel. Under the guidance of sampling parameter $\Theta$, a geometrically deformed feature map $F^{'}_{c}$ is obtained from $\Theta$ and $F_c$ based on deformable convolution ${f_{DC}(\cdot)}$:
\begin{equation}
F^{'}_{c} = f_{DC}(F_c, \Theta).
\end{equation}
Specifically, for each position $p$ on the output $F^{'}_{c}$, the deformable convolution $f_{DC}(\cdot)$ is applied as follow:
\begin{equation}
F^{'}_{c}(p) = \sum_{k=1}^{\mathcal{R}}w(p_k)\cdot x(p + p_k + \Delta p_k) \cdot \Delta m_k,
\end{equation}
where the $w(p_k)$ indicates the weight of the deformable convolution kernel at $k$-th location. The convolution is operated on the irregular positions ($p_k$ + $\Delta$ $p_k$) where $\Delta$ $p_k$ may be fractional.  Followed \cite{Deformablev1}, the operation is implemented by using bilinear interpolation. At last, the output of the FDSC module is fed to the mixer and  $F^{'}_{c}$ is then concatenated with $F_s$ like a commonly used skip connection  \cite{DBLP:conf/miccai/RonnebergerFB15}.

Deformable convolution introduces 2D offsets to the regular grid sampling locations in the standard convolution. It enables free-form deformation of the sampling grid. There are lots of areas of the same color in character images, such as background color and character color. By using deformable convolution, an area can be related to any other area with the same color. It is difficult to optimize the non-unique solution. To efficiently use our FDSC module, we impose a constraint on the offsets $\Delta p$. We introduce the constraint in detail in Subsection \ref{sec:loss}, and  demonstrate the visualization of the offsets $\Delta p$ in Section \ref{visualization}.

Our FDSC module aims to deform the spatial structure of the content image in the feature space. It is crucial to select which level of features to be transformed, as low-level features contain more structure and spatial information than high-level features. Experiments in Section \ref{sec:FDSC_analysis} demonstrate the analysis of the performance of the model with different numbers of the FDSC module.

 \textbf{FDSC-attn.} 
In the inner skip connection, we integrate the local spatial attention model, aiming to explore local relationships by predicting weights at a position in regard to the features at its neighboring positions. The detail of FDSC-attn is shown in Figure \ref{fig:2}(c). The deformable feature $F^{'}_c \in \mathbb{R}^{h \times w \times c}$ produced by FDSC is first projected to the latent space as query, key, value $Q(F^{'}_c), K(F^{'}_c), V(F^{'}_c) \in \mathbb{R}^{h\times w \times c}$ by using $1 \times 1$ convolution layers. For each location $(i, j)$ within the spatial dimensions, we extract a patch with size $s$ from $K$ centered at $(i,j)$, denoted as $\bs{k} \in \mathbb{R}^{s \times s \times c}$. Then the weight $\bs{w} \in \mathbb{R}^{s \times s \times c}$ is obtained by reshaping the estimation of a feed-forward network (FFN):
\begin{equation}
     \bs{w}=reshape(FFN(concat(flatten(\bs{k}),\bs{q})).
\end{equation}
 The FFN is computed by giving the concatenation of the flattened patch query $flatten(k)$ and its corresponding query vector $\bs{q} \in \mathbb{R}^{1 \times 1 \times c}$ at $(i,j)$. The FFN consists of a fully-connected (FC) layer followed by leaky ReLu and a linear FC layer. Here $w$ plays a conceptually equivalent role as the softmax attention map of the traditional key query aggregation [83, 98, 65]. The output vector $\boldsymbol{o} \in \mathbb{R}^{1 \times 1 \times c}$ at $(i,j)$ is then calculated by element-wise multiplication of the predicted $w$ and the value patch with the same size from $V$ centered at $(i,j)$, denoted as $\bs{v} \in \mathbb{R}^{s \times s\times c}$: $\bs{o}(i,j) = \bs{w} \odot \bs{v}$. We loop over all the $(i, j)$ to constitute an output $O_{a}$ as the attention output. Different from deformable convolution, the additional attention module provides a heat map between the feature of a certain location and its context patch. It allows our model to look over the local attention information, then generates a 'soft' deformed feature according to its current position feature and context. 
\begin{figure}[t]
\center
   \includegraphics[width=0.9\linewidth]{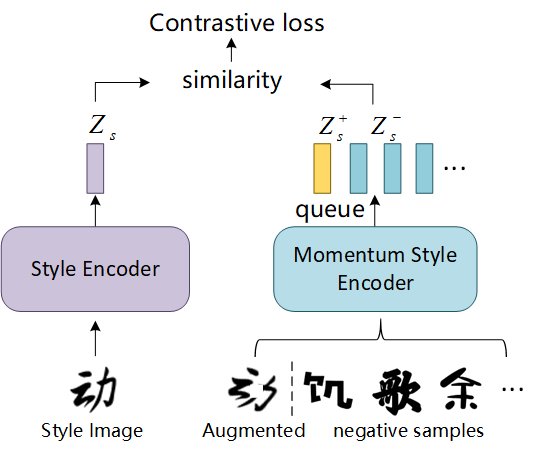}
   \caption{Illustration of our contrastive representation learning. The model employs the style reference image and its augmented image as a positive pair. The other image samples of the dataset are negative samples.}
\label{fig:contra}
\end{figure}
\subsection{Unsupervised Feature Representation for Style Encoder}
To learn a robust representation for a given font, we hope to learn consistent style features regardless of its content. To this end, we define several data augmentation operations to construct positive counterparts for fonts. Then we adopt contrastive learning to force the positive pairs to be similar in the latent space. The contrastive representation learns the visual styles of images by maximizing the mutual information between an image and its augmented version in contrast to other negative images within the dataset.

A detailed illustration of the proposed contrastive representation learning is shown in Figure \ref{fig:contra}. Our style encoder aims to learn a robust map of the style image $I_s$ to style code $Z_s$. Similar to MOCO \cite{DBLP:conf/cvpr/He0WXG20}, a momentum style encoder is used to map the positive counterpart $I_s^{+}$ as positive codes $Z_s^{+}$. We construct a queue of data samples that are stored from the previously sampled images and mapped them as negative codes $Z_s^{i-}$. Then we use a contrastive loss to unitize the similarity of the positive pair $(Z_s,Z_s^{+})$ and the dissimilarity of negative pairs:$(Z_s,Z_s^{i-})$.
\begin{equation}\label{eq:sty}
     \mathcal{L}_{sty}=-\log \frac{\exp (Z_s\cdot Z_s^{+}/\tau)}{\exp (Z_s \cdot Z_s^{+}/\tau) + \sum_{i=1}^{N}\exp (Z_s\cdot Z_s^{i-}/\tau)},
\end{equation}
where $\tau$ is a temperature hyper-parameter \cite{wu2018unsupervised}. The sum is over one positive and $N$ negative sample. This loss is the log loss of an (N+1)-way softmax-based classifier aiming to classify $Z_s$ as $Z_s^{+}$. The momentum style encoder is updated followed by \cite{DBLP:conf/cvpr/He0WXG20}.
The contrastive representation facilitates the style encoder to learn robust style features, which alleviates the impact of using different content images of a certain style. Experiments in Section \ref{sec:robustness} demonstrate the influence by varying the number of the reference style images.

\subsection{Loss Function}\label{sec:loss}
Our model aims to achieve automatic font generation via an unsupervised method. In addition to the contrastive loss in Eq. \ref{eq:sty}, we adopt four losses: 1) adversarial loss is used to produce realistic images; 2)  content consistent loss is introduced to encourage the content of the generated image to be consistent with the content image; 3) image reconstruction loss is used to maintain the domain-invariant features; 4) deformation offset normalization is designed to prevent excessive offsets of the FDSC module. We introduce the formula of each loss and the full objective in this section.

\textbf{Adversarial loss: } When character images are generated from the generative network, \textbf{a multi-task discriminator} is adopted to conduct discrimination for each style simultaneously. For each style, the output of the discriminator is a binary classification whether the input image is a real image or a generated image. As there are several different styles of fonts in the training set, the discriminator outputs a binary vector whose length is the number of styles. In all, our model aims to generate plausible images by solving a mini-max optimization problem. The generative network $G$ tries to fool discriminator $D$ by generating fake images. The adversarial loss penalty is the wrong judgment when real/generated images are input to the discriminator.
\begin{equation}
\begin{aligned}
\mathcal{L}_{adv} = \max_{D_s}\min_{G} \mathbb{E}_{I_s \in P_s, I_c \in P_c} [ \log D_s(I_s)
\\
+\log (1-D_s(G(I_s, I_c)))],
\end{aligned}
\end{equation}
where $D_s$($\cdot$) denotes the logit from the corresponding style of the discriminator’s output.

\textbf{Content consistent loss: }adversarial loss is adopted to help the model to generate a realistic style while ignoring the correctness of the content. To prevent mode collapse and ensure that the features extracted from the same content can be content consistent after the content encoder $f_c$, we impose a content consistent loss here:
\begin{equation}
\label{Lcnt}
\mathcal{L}_{cnt} =  \mathbb{E}_{I_s \in P_s, I_c \in P_c}\left\|Z_c - f_c(G(I_s, I_c))\right\|_1 .
\end{equation}
${L}_{cnt}$ ensures that given a source content image $I_c$ and corresponding generated images, their feature maps are consistent after content encoder $f_c$.

\textbf{Image reconstruction loss: } To ensure that the generator can reconstruct the source image $I_c$ when given its origin style, we impose a reconstruction loss:
\begin{equation}
\label{Limg}
\mathcal{L}_{img} =  \mathbb{E}_{I_c \in P_c}\left\|I_c - G(I_c,I_c)\right\|_1 .
\end{equation}
The objective helps preserve domain-invariant characteristics (\eg content) of its input image $I_c$.

\textbf{Deformation offset normalization: }
The deformable offsets enable free-form deformation of the sampling grid.
As there are lots of areas of the same color between input images and generated images (such as background color and character color), it leads to a non-unique solution that is difficult to optimize. Meanwhile, the font generation focus on the stroke relationship between the content character image and the target character image, such as the thickness and tips of the stroke. However, given images with the same content but different style, the position of the same stroke in these two images are close. To efficiently use this deformable convolutional network, we impose a constrain on the offsets $\Delta p$:
\begin{equation}
\label{offsetloss}
\mathcal{L}_{offset} =  \frac{1}{|\mathcal{R}|}\left\|\Delta p\right\|_1,
\end{equation}
where $\Delta p$ denotes offsets of the deformable convolution kernel, $|\mathcal{R}|$ denotes the number of the convolution kernel. 

\textbf{Overall objective loss: }Combining all the above-mentioned losses, we have the overall loss function for training our proposed framework: 
\begin{equation}
\begin{aligned}
\mathcal{L}& = \mathcal{L}_{adv} + \lambda_{img}\mathcal{L}_{img} + \lambda_{cnt}\mathcal{L}_{cnt} \\&+ \lambda_{sty}\mathcal{L}_{sty} + \lambda_{offset}\sum^{N}_{i}\mathcal{L}_{offset}/N,
\end{aligned}
\end{equation}
where $\lambda_{adv}$, $\lambda_{img}$, $\lambda_{cnt}$, $\lambda_{sty}$, and $\lambda_{offset}$ are hyper-parameters to control the weight of each loss, and $N$ indicates the number of FDSC modules.  In our model, the generative network aims to minimize the overall object loss, while the discriminator aims to maximize it.

\section{Experiments}
In this section, we evaluate our proposed model for the Chinese font generation task. We first introduce the implementation details of our experiments. Extensive experiments demonstrate the superiority of our model. We also provide ablation studies on the effects of FDSC, FDSC-attn, and objective function.

\subsection{Implementation Detail}
\subsubsection{Network Architecture}\label{sec:arc}
The generative network consists of a style encoder, a content encoder, a mixer, and two FDSC modules. The style encoder is composed of 7 convolution layers and a fully connected (FC) layer. The convolution layer is of $3 \times 3$ kernel with nonlinear activation of ReLU. The FC layer maps the style feature maps into a style representation vector of 128 dimensions.
The architecture of the content encoder and mixer is symmetrical. The detailed architectures of the style encoder, content encoder, and mixer are shown in Table \ref{architecture1}. 
The two FDSC modules consist of an FDSC and an FDSC-attn module. The FDSC is constructed by a deformable convolution layer with $3 \times 3$ kernel. FDSC-attn consists of a deformable convolution layer and a local spatial attention module with patch size $s=13 \times 13$. The discriminator contains several residual blocks and three convolution layers. The detailed information on the discriminator is listed in Table \ref{architecture2}. FRN indicates filter response normalization \cite{singh2020filter}.
\subsubsection{Training Strategy}
We initial the weights of convolutional layers with He initialization \cite{DBLP:conf/iccv/HeZRS15}, in which all biases are set to zero and the weights of linear layers are sampled from $N(0, 0.01)$. We use Adam optimizer with $\beta_1$ = 0.9 and $\beta_2$ = 0.99 for style encoder, and RMSprop optimizer with $\alpha$ = 0.99 for the content encoder and mixer. We train the whole framework with 200,000 iterations and the learning rate is set to 0.0001 with a weight decay of 0.001. 
We train the model with a hinge version adversarial loss \cite{DBLP:journals/corr/TranRB17} with R1 regularization \cite{DBLP:conf/icml/MeschederGN18} using $\gamma=10$. We optimize our DGFont++ with  $\lambda_{img} = 0.1$, $\lambda_{cnt} = 0.1$, $\lambda_{sty} = 0.1$,  $\lambda_{offset} = 0.1$. 
The temperature hyper-parameter in Eq. \ref{eq:sty} is set as $\tau=0.07$ by default \cite{DBLP:conf/cvpr/He0WXG20}. In the testing process, we use ten reference images to compute an average style code for the generation process.

\subsubsection{Data augmentation}
 We use several data augmentation operations of spatial and geometric transformation to construct positive counterparts for fonts.  
Specifically, for a given style reference image, we conduct the following data augmentation operations successively: randomly scaling (from 0.7 to 1.2), rotation (degree from -20 to 20), horizontal and vertical translation (shift value range from 0 to 0.2 of image size) and random erasing with filled 255.
\begin{table}[th]
\center
\scalebox{0.95}{
\begin{tabular}{cccccc}
\toprule
& Operation & Kernel & Resample & Padding & Features \\ \midrule
\multirow{10}{*}{\rotatebox{90}{Style encoder}} & Convolution                            & 3           & MaxPool  & 1       & 64           \\
                                 & Convolution                            & 3           & MaxPool  & 1       & 128          \\
                                 & Convolution                            & 3           & -        & 1       & 256          \\
                                 & Convolution                            & 3           & MaxPool  & 1       & 256          \\
                                 & Convolution                            & 3           & -        & 1       & 512          \\
                                 & Convolution                            & 3           & MaxPool  & 1       & 512          \\
                                 & Convolution                            & 3           & -        & 1       & 512          \\
                                 & Convolution                            & 3           & MaxPool  & 1       & 512          \\
                                 & Avg pooling                        & -           & -        & -       & 512          \\
                                 & FC & -           & -        & -       & 128          \\ \hline
\multirow{5}{*}{\rotatebox{90}{\begin{tabular}[c]{@{}l@{}}Content\\ encoder\end{tabular}}} & Deform. conv.                        & 3           & -        & 1       & 32           \\
                                 & Deform. conv.                        & 4           & stride-2 & 1       & 64          \\
                                 & Deform. conv.                        & 4           & stride-2 & 1       & 128          \\
                                 & Convolution                        & 4           & stride-2 & 1       & 256          \\
                                 & Res block & 3           & -        & 1       & 256          \\
                                 
                                 & Res block & 3           & -        & 1       & 256          \\\hline
\multirow{5}{*}{\rotatebox{90}{Mixer}}           & Res block & 3           & -        & 1       & 256          \\
   & Res block& 3           & -        & 1       & 256          \\
                                 & Convolution                            & 5           & Upsample & 2       & 128          \\
                                 & Convolution                            & 5           & Upsample & 2       & 64          \\
                                 & Convolution                            & 5           & Upsample & 2       & 32           \\
                                 & Convolution                            & 3           & -        & 1       & 3            \\ \bottomrule
\end{tabular}
}
\caption{Architecture of generative network}
\label{architecture1}
\end{table}

\begin{table}[th]
    \center
    \begin{tabular}{ccccc}
    \toprule
   Operation      & Kernel  & Resample  & Features & Normalization  \\ 
   \midrule
   Convolution    & 3  & -     & 64   & -   \\
      Res block & 3 & -  & 64   & FRN    \\
      Res block & 3           & AvgPool  & 128          & FRN         \\
     Res block & 3           & -   & 128          & FRN              \\
      Res block & 3           & AvgPool   & 256          & FRN   \\
      Res block & 3           & -  & 256          & FRN     \\
      Res block & 3           & AvgPool & 512          & FRN      \\
      Res block & 3           & -  & 512          & FRN    \\
      Res block & 3           & AvgPool   & 1024         & FRN      \\
      LeakyReLU      & -           & -        & -       & - \\
      Convolution    & 4           & -  & 1024         & -          \\
      LeakyReLU      & -           & -        & -       & - \\
      Convolution    & 1   & AvgPool & 221          & -  \\ \midrule
    \multicolumn{5}{c}{AvgPool: Average Pooling; \ \   LeakyReLU: Slope 0.2.} \\\bottomrule
     \specialrule{0em}{0pt}{1.5pt}
    \end{tabular}
    \caption{Architecture of discriminative network}
    \label{architecture2}
\end{table}

\subsubsection{Compared Methods}
We compare our model with the following methods for Chinese font generation: 
\begin{table*}[t]
\begin{center}
\setlength{\tabcolsep}{10pt}{
\begin{tabular*}{15.5cm}{lclccccc}
\toprule[1.0pt]
Methods & one-to-many& training & L1 loss$\downarrow$& RMSE$\downarrow$& SSIM$\uparrow$& LPIPS$\downarrow$ & FID$\downarrow$\\
\midrule[1.0pt]
\multicolumn{8}{c}{\textbf{Seen} fonts} \\
\midrule[1.0pt]
EMD \cite{EMD} - 80  & $\checkmark$ & paired &0.0538 &0.1955 &0.7676&0.1036 &89.65\\
Zi2zi \cite{zi-2-zi}  - 80 & $\times$ & paired &\textbf{0.0521} &0.1802 &0.7789 & 0.1065 &142.23\\
Cycle-GAN \cite{Zhu2017UnpairedIT}  - 80 & $\times$ & unpaired &0.0863 &0.2555 &0.6392 &0.1825 &175.24\\
GANimorph \cite{ganimorph}  - 80 & $\times$ & unpaired &0.0563 &\textbf{0.1759} &\textbf{0.7808} &0.1403 &72.89\\
FUNIT \cite{DBLP:conf/iccv/0001HMKALK19}  - 80 & $\checkmark$ & unpaired & 0.0807 &0.2510 &0.6669 &0.1216 &53.77\\
DGFont - 80 & $\checkmark$ & unpaired &0.0562 &0.1994 &0.7580 &\textbf{0.0814} & 46.15\\
DGFont++  - 80 & $\checkmark$&unpaired &0.0580&0.2032&0.7502&0.0875&\textbf{43.56}\\
\midrule[0.2pt]
FUNIT \cite{DBLP:conf/iccv/0001HMKALK19}  - 128  &$\checkmark$&unpaired&0.0894&0.2766&0.7193&0.2113&36.40\\
MXFont \cite{Park_2021_ICCV}- 128 &$\checkmark$&paired&0.0704&0.2353&0.7622&0.1505&46.69\\
DGFont  - 128 &$\checkmark$&unpaired&0.0584&0.2166&0.7871&0.1204&41.69\\
DGFont++ - 128 &$\checkmark$&unpaired&\textbf{0.0567}&\textbf{0.2131}&\textbf{0.7910}&\textbf{0.1151}&\textbf{35.83}\\
\midrule[1.0pt]
\multicolumn{8}{c}{\textbf{Unseen} fonts} \\
\midrule[1.0pt]
EMD \cite{EMD} - 80 & $\checkmark$ & paired & 0.0430 &0.1755 &0.7849 &0.1255 &82.53\\
FUNIT \cite{DBLP:conf/iccv/0001HMKALK19} - 80 & $\checkmark$ & unpaired &0.0588 &0.2089 &0.7417 &0.1125 &59.98\\
DGFont - 80 & $\checkmark$ & unpaired & \textbf{0.0414} &\textbf{0.1709} &\textbf{0.7982}& \textbf{0.0867}& \textbf{50.29}\\
\midrule[0.2pt]
FUNIT \cite{DBLP:conf/iccv/0001HMKALK19}  - 128  &$\checkmark$&unpaired&0.0703&0.2426&0.7655&0.1734&35.76\\
MXFont \cite{Park_2021_ICCV} - 128 &$\checkmark$&paired&0.0561&0.2057&0.8121&0.1269&39.61\\
DGFont - 128&$\checkmark$ & unpaired &\textbf{0.0517}&\textbf{0.2032}&\textbf{0.8146}&0.1250&41.59\\
DGFont++ - 128 & $\checkmark$&unpaired&0.0532&0.2059&0.8125&\textbf{0.1243}&\textbf{34.54}\\
\bottomrule[1.0pt]
\end{tabular*}
}
\end{center}
\caption{\textbf{Quantitative evaluation on the whole dataset}. We evaluate the methods on seen and unseen font sets. The bold number indicates the best. 80 and 128 in the first column indicate the training and testing image size.}
\label{table1}
\end{table*}
\begin{itemize}
\item Cycle-GAN \cite{Zhu2017UnpairedIT}: Cycle-GAN is an unsupervised image-to-image translation method, which consists of two generative networks which translate images from one domain to another using a cycle consistency loss. 
\item EMD \cite{EMD}: EMD is a supervised font generation method that is optimized by L1 distance loss between ground-truth and generated images. It employs an encoder-decoder architecture and separates style/content representations. 
\item Zi2zi \cite{zi-2-zi}: Zi2zi is a supervised font generation method based on pix2pix \cite{Isola2017ImagetoImageTW}, it achieves font generation and uses Gaussian Noise as category embedding to achieve multi-style transfer.
\item GANimorph \cite{ganimorph}: GANimorph is an unsupervised image-to-image translation method for  representing shape deformation by introducing a discriminator with dilated convolutions to get a more context-aware generator.
\item FUNIT \cite{DBLP:conf/iccv/0001HMKALK19}: FUNIT is a few-shot unsupervised image-to-image translation model which separates content and style of natural animal images and combines them with adaptive instance normalization (AdaIN) layer.
\item  MXFONT~\cite{Park_2021_ICCV}: MXFont is the state-of-the-art few-shot font generation method with multiple localized experts by utilizing decomposable glyphs.
\end{itemize}
\begin{figure*}
\begin{center}
\subfigure[Easy Cases (\eg~printing typeface).]{
\begin{minipage}[b]{1\textwidth}
\includegraphics[width=0.99\textwidth]{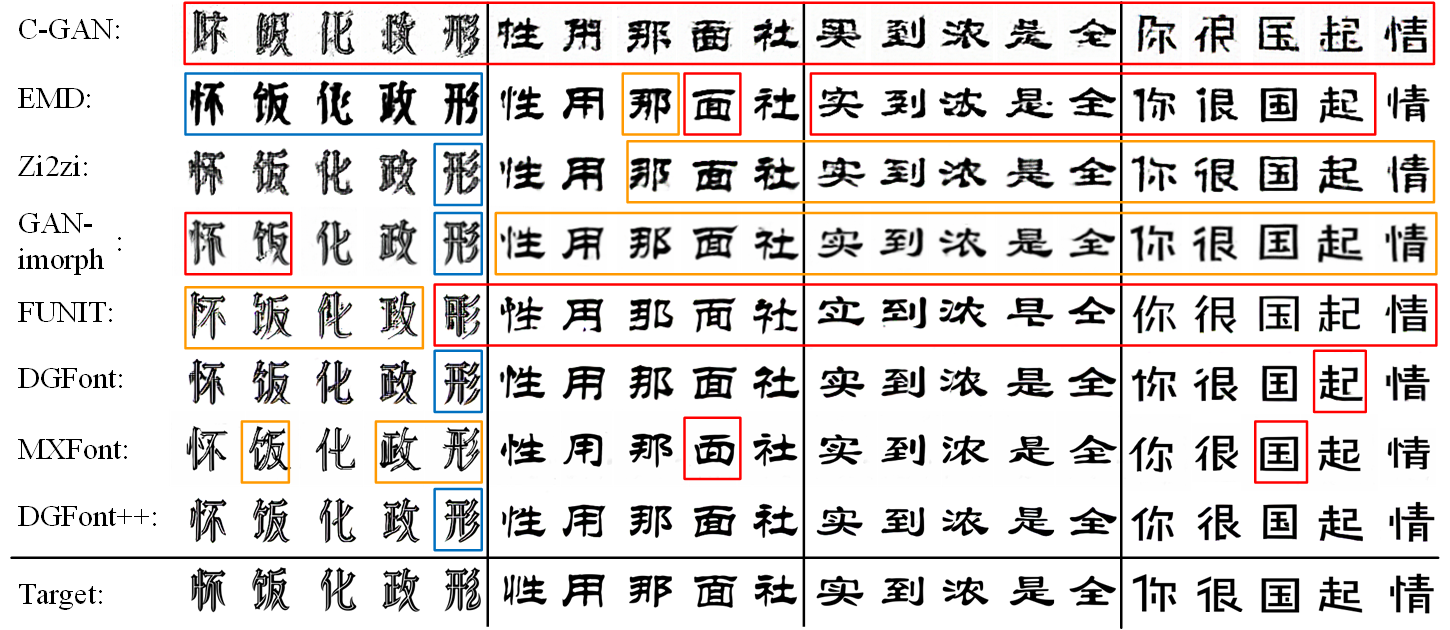}\label{fig:4a}
\end{minipage}
}
\subfigure[Challenging Cases~(\eg~calligraphy, wordart).]{
\begin{minipage}[b]{1\textwidth}
\includegraphics[width=0.99\textwidth]{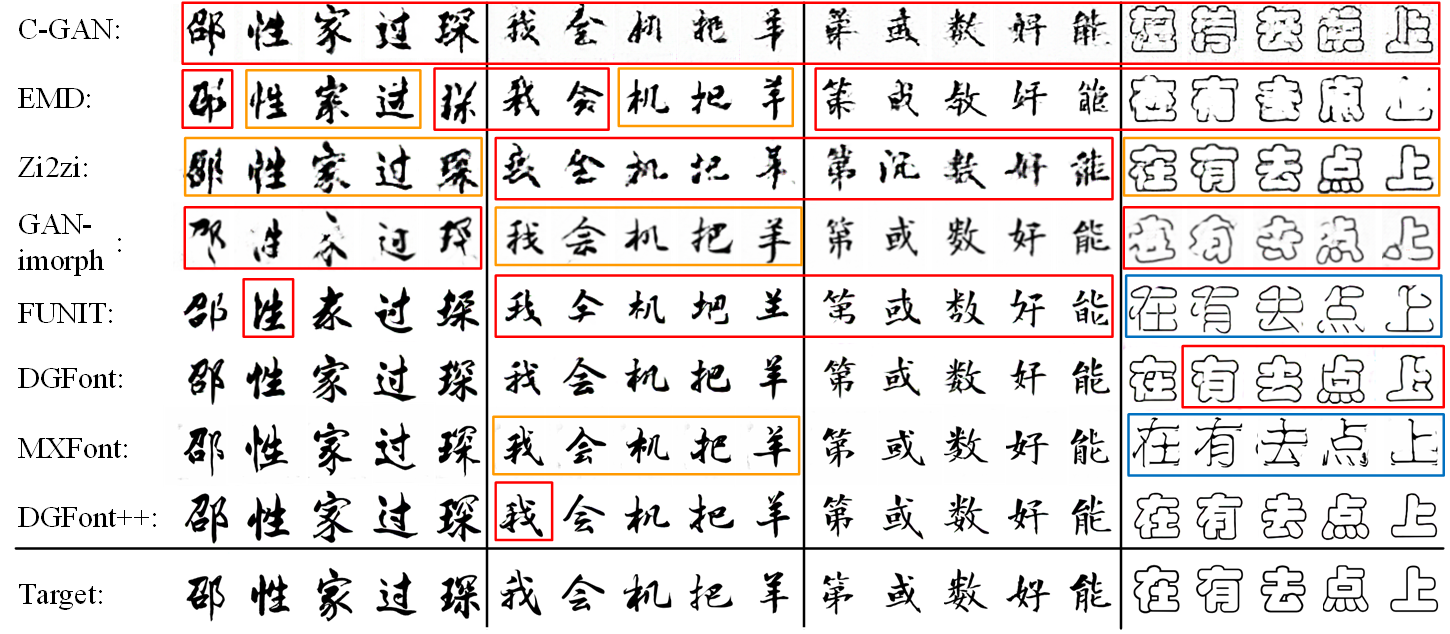}\label{fig:4b}
\end{minipage}
}
\end{center}
\caption{Comparisons between our model and the state-of-the-art methods. The red boxes highlight failures of structure preservation, the blue boxes highlight failures of style transfer, and the orange boxes highlight the blur and noisy outputs.} \label{fig:results}
\label{reuslts}
\end{figure*}
\subsubsection{Dataset and Evaluation Metrics}
We collect a dataset containing 231 fonts (styles) including printing and handwriting fonts, each of which has 1143 commonly used Chinese characters (content). The dataset is randomly partitioned into a training set and a testing set. The training set contains 221 fonts, and each font contains 800 characters. The testing set consists of two parts. One testing part is the remaining 343 characters of the 221  fonts. Another part is the remaining 10 fonts for testing the generalization for unseen fonts. The image size is $128 \times 128$, which is consistent with most mainstream methods \cite{DBLP:conf/iccv/0001HMKALK19,Park_2021_ICCV}. In order to compare with the latest font generation methods (\ie MXFont \cite{Park_2021_ICCV}), the collected characters can be decomposed into sub-characters. Differently, the dataset in our conference version is of $80 \times 80$ resolution and some of the characters in the conference version cannot be decomposed.
For a fair comparison, all the experiments on $128 \times 128$ images use the newly collected dataset, and experiments on $80\times 80$ images use the dataset of the conference version.

We employ five metrics L1 loss, Root Mean Square Error (RMSE), SSIM, LPIPS \cite{DBLP:conf/cvpr/ZhangIESW18} and FID \cite{DBLP:conf/nips/HeuselRUNH17} for evaluation. L1 loss, RSME, and Structural Similarity (SSIM) is widely used in font generation task with known ground truths \cite{EMD}. L1 loss calculates the L1-norm of the distance between generated image and ground-truth. RMSE utilizes the mean square error to give an overall evaluation. SSIM  calculates the global mean and variance to assess the structural similarity. 
Meanwhile, Learned Perceptual Image Patch Similarity (LPIPS) is another metric to compute the distance between the generations and ground truths in the perceptual domain. Besides, Fréchet Inception Distance (FID) is employed to measure the realism of the generated images. The FID score is calculated between the distributions of the generated and real data for each style.

\subsection{Comparison with State-of-art Methods}
\begin{figure}[t]
\begin{center}
   \includegraphics[width=1\linewidth]{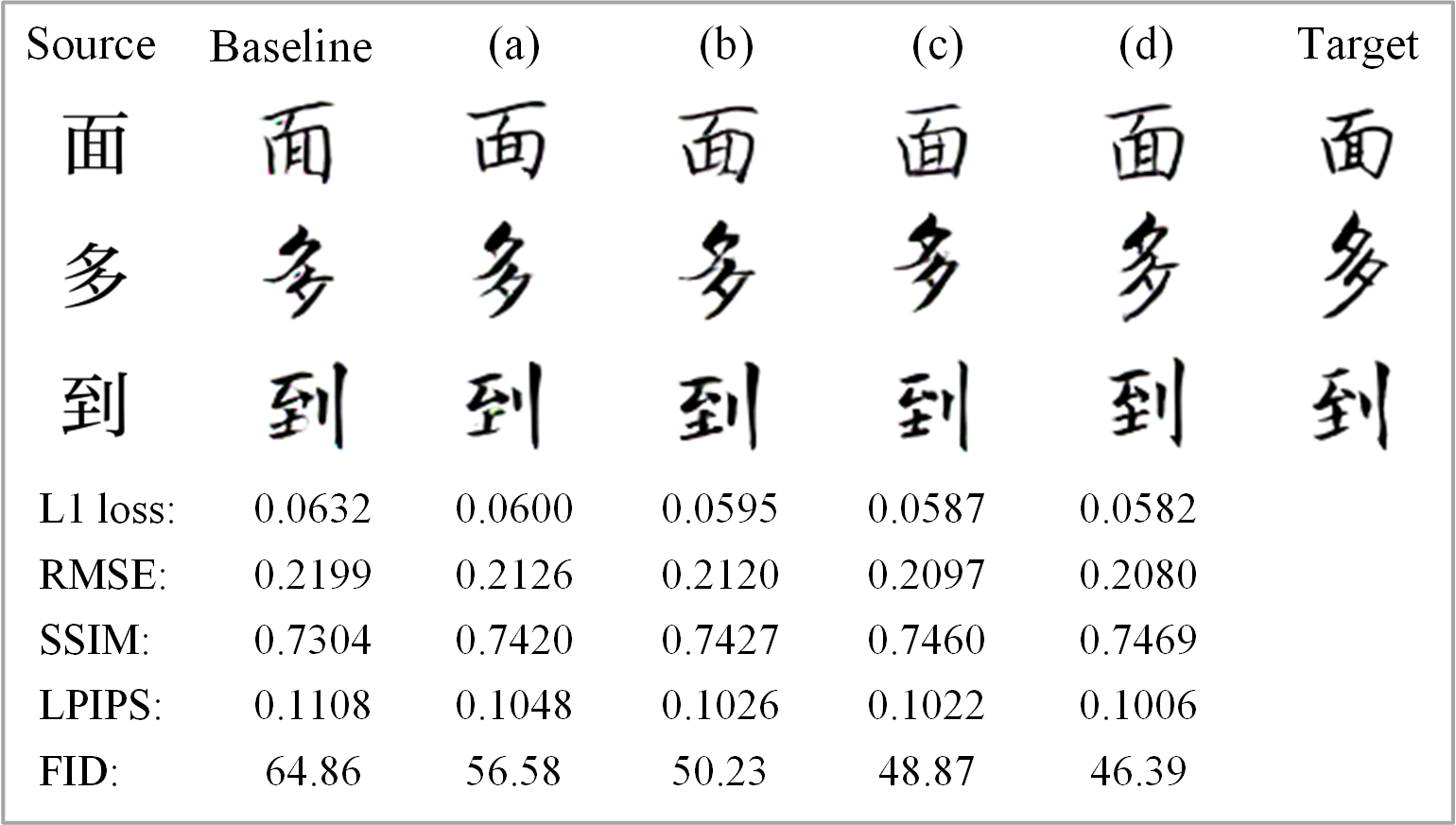}
\end{center}
   \caption{\textbf{Effect of different components in DGFont}. We add different parts into our baseline successively. (a) Replace the first three convolution layers of content encoder with deformable convolution layers; (b) add one FDSC module (without normalization); (c) impose normalization on the FDSC module; (d) add another FDSC module (\ie DGFont).}
\label{fig:ablation}
\end{figure}

\subsubsection{Quantitative comparison}
The quantitative results are shown in Table \ref{table1}. In addition to comparing methods of generating 80 $\times$ 80 images as our previous version \cite{xie2021dg}, we adapt our model to generate 128 $\times 128$ and compare with recent work \cite{Park_2021_ICCV} designed for the image of 128 $\times$ 128. For the one-to-one image translation models (\ie~Cycle-GAN and GANimorph), we train models individually for each target style. In the experiment of generating images of 80 $\times$ 80 size, results show that our methods (\ie~DGFont and DGFont++) are comparable to the compared methods in pixel-level evaluation metrics, \eg~L1 loss, RMSE, SSIM. It is noted that these metrics focus on the pixel-wise comparison between generated image and ground-truth while ignoring the feature similarity. Pixel-wise metrics tend to predict high scores for blur images that are not consistent with human perception \cite{Johnson2016PerceptualLF}. In perceptual-level metrics (\ie~FID \cite{DBLP:conf/nips/HeuselRUNH17} and LPIPS \cite{DBLP:conf/cvpr/ZhangIESW18}), our methods outperform all the compared methods for both seen font and unseen font.

 In the experiment of generating images of 128 $\times$ 128 sizes, we compare our method to FUNIT and MXFont. We observe MXFont achieves a better score than FUNIT in terms of L1 loss, RMSE, SSIM, and LPIPS, while FUNIT outperforms MXFont in terms of FID score. It is because MXFont uses pixel-wise loss to supervise the training of the model. FUNIT has a higher score in FID which measures the realism and quality of generated images. In contrast, our methods (\ie~DGFont and DGFont++) outperform FUNIT and MXFont in all metrics and achieve state-of-the-art performance. Our DGFont++ has achieved better performance than DGFont. Detailed analysis for DGFont and DGFont++ is discussed in Section \ref{sec:dgfont++}. To explore the generality of our model, we also generate imitations for unseen styles whose results are shown in Table  \ref{table:dgfont++}. It demonstrates that our model outperforms all the compared methods, including the few-shot font generation method, MXFont.

\subsubsection{Qualitative comparison}
Figure \ref{reuslts} demonstrates the qualitative comparison between our method and the state-of-the-art methods.
To explore the capability of deforming and transforming source character patterns (\eg~stroke, skeleton), we display the visual comparisons as two cases: easy cases and challenge cases. Characters in easy cases are close to printing typefaces that do not have cursive writing, while in challenge cases characters are of WordArt and calligraphy fonts that are hollow or have joined-up writing.
We observe that the results of Cycle-GAN can hardly recognize and tend to be noisy. In easy cases, characters generated by Zi2zi, EMD, and GANimorph can maintain a complete structure, but they are usually vague. In challenging cases, they can only generate parts of characters or sometimes unreasonable structures. 
FUNIT can generate characters with a clear background but the generated characters lose their structure to some degree. In the challenging cases (Figure \ref{fig:4b}), FUNIT generates characters of incomplete structure when the target font is the cursive writing style. In the case of generating hollow font, FUNIT fails to generate the character of reference style (see the last column in Figure \ref{fig:4b}).
MXFont is the state-of-the-art few-shot font generation method. In most cases, MXFont and our method generate images of comparable quality. Notably, MXFont requires paired datasets while our methods are in an unsupervised manner.
Also, we observe that MXFont fails in generating hollow font. In contrast, our proposed methods, DGFont, and DGFont++ can not only generate characters with complete structure but also learn joined-up writing. Also, DGFont++ can produce more clear and more complete images than DGFont. More ablation study for DGFont and DGFont++ is discussed in the next section. 

\subsection{Ablation Study}\label{ablation}
In this part, we conduct ablation studies for our model. We set our baseline model as a normal encoder-decoder model which consists of a style encoder, content encoder, and mixer, but no deformable convolution layer, FDSC module, and contrastive loss. In Section \ref{sec:abl_dgfont}, we conduct ablation studies for incremental modules and losses between DGFont and the baseline model. In Section \ref{sec:dgfont++}, we conduct ablation studies for incremental  modules and losses between DGFont and DGFont++.
\subsubsection{Baseline model vs. DGFont}\label{sec:abl_dgfont}
We add deformable convolution, feature deformation skip connection, and deformable offset normalization successively on the baseline model and get the full model of DGFont. The experiments are conducted in challenging cases to explore the functionality of each component for DGFont. Qualitative and quantitative comparisons are shown in Figure \ref{fig:ablation}. 

1) \textbf{Effectiveness of deformable convolution in the content encoder.} Figure \ref{fig:ablation}(a) shows the results by replacing the first three convolution layers of the content encoder with deformable convolution layers. We can see that the quantitative results improve obviously in terms of L1 loss, RMSE, and SSIM. This indicates that deformable convolution layers in the content encoder effectively help improve the performance of our model.

2) \textbf{The influence of the FDSC module.} In this part, we add an FDSC module (without offset normalization in Eq. \ref{offsetloss}) that connects the features after the first layer and penultimate layer. Results are shown in Figure \ref{fig:ablation}(b). Comparing with Figure \ref{fig:ablation}(a), we observe that the generated characters preserve more structure information and are able to reconstruct the complete structure of characters. 

3) \textbf{Effectiveness of deformable offset constraint.} We investigate the impact of deformable offset normalization by comparing FDSC modules without and with offset normalization. As shown in Figure \ref{fig:ablation}(b) and (c), adding offset normalization helps the model generate images whose style becomes more similar to the target.

4) \textbf{Effectiveness of two FDSC modules.} Figure \ref{fig:ablation}(d) shows the results of full DGFont with two FDSC modules. It is noted that the generated images get more details, less noise, and achieve better quantitative results.
\subsubsection{DGFont vs. DGFont++}\label{sec:dgfont++}
\begin{figure}[t]
\center
\includegraphics[width=\linewidth]{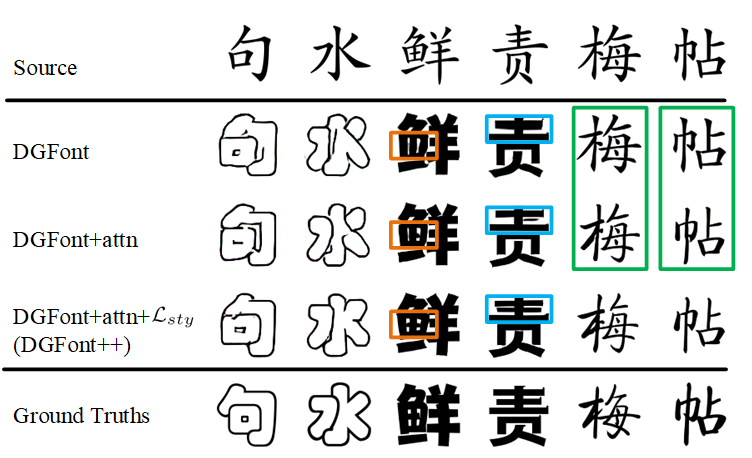}
\caption{Visual samples of ablation study for DGFont++.}
\label{fig:dgfont++}
\end{figure}

Based on DGFont, we then replace FDSC as FDSC-attn in the inner skip connection (indicated as DGFont+attn), and successively add contrastive learning (indicated as DGFont+attn+$\mathcal{L}_{sty}$) to get the full model of DGFont++. Results of ablation study for DGFont++ (\ie DGFont+attn+$\mathcal{L}_{sty}$) are shown in Table \ref{table:dgfont++} and Figure \ref{fig:dgfont++}. Experiments are conducted on the image of size 128 $\times$ 128. We analyze the results and effectiveness as follows.

1) \textbf{Effectiveness of the FDSC-attn module.}
In this part, we investigate the effectiveness of  FDSC-attn compared with FDSC. Quantitative results are shown in the second row of Table \ref{table:dgfont++}(indicated as `` DGFont+attn"). DGFont+attn achieves a close score in pixel-wise metrics (\ie~L1 loss, RMSE, SSIM) while outperforming DGFont in terms of perceptual metrics (\ie, FID and LPIPS). The qualitative results for FDSC-attn are shown in Figure \ref{fig:dgfont++}. Compared to DGFont, the generated images learn the target style well. For example, in the fourth and fifth columns, DGFont fails to transfer style whose results remain the style as the source style, while the model with FDSC-attn succeeds in generating images of the target style.

2) \textbf{Effectiveness of contrastive loss.}
We investigate the impact of style contrastive loss. In Table \ref{table:dgfont++}, we observe that the model with contrastive loss produces results of the highest score in all metrics. The fourth row of Figure \ref{fig:dgfont++} (indicated as ``DGFont+attn+$\mathcal{L}_{sty}$") shows the qualitative results for contrastive loss. Results show that our DGFont++ is able to generate high-quality images without missing strokes or noisy points. It is because the introduced contrastive learning helps to learn robust and content-invariant style representation, which helps to maintain complete structure and strokes of generated glyph images. More analysis for contrastive representation learning is demonstrated in the next section.

\begin{figure*}[t]
\center
\includegraphics[width=\linewidth]{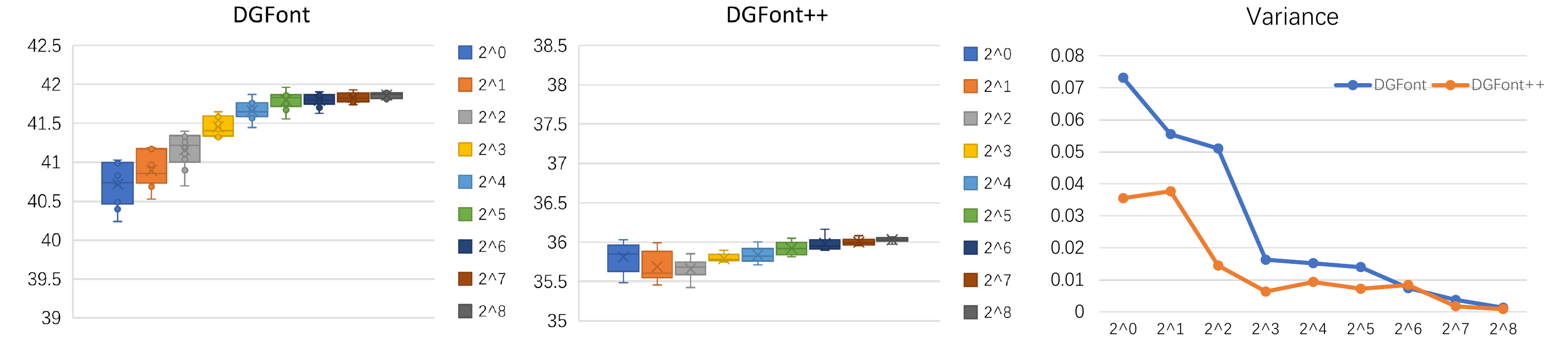}
\caption{Performance changes by varying the size of the reference set. We report how the performance is affected by the size of the reference style images. For each case, we calculate the FID score by random sampling the style images ten times. Figures (a) and (b) show the statistics of the FID score as box plots. Figure (c) shows the variance of FID for each model.}
\label{fig:robustness}
\end{figure*}

3) \textbf{Robustness Analysis.}\label{sec:robustness}
In this subsection, we investigate the robustness of our model by varying the size of the style reference images. We report the FID score of DGFont++ and DGFont by varying the size of the reference set from 1 ($2^0$) to 256 ($2^8$). Results are shown in Figure \ref{fig:robustness}. For each case, we evaluate the performance by randomly sampling the style images ten times, and the results are demonstrated as box plots. We use the same y-axis interval in Figure \ref{fig:robustness}(a) and Figure \ref{fig:robustness}(b) for better comparison. Results show that DGFont++ achieves better FID scores, whose results also have a smaller fluctuation under ten times of inference with the same number of references. Meanwhile, Figure \ref{fig:robustness}(c) shows the variance of FID scores for DGFont++ and DGFont, which demonstrates that DGFont++ has a lower variance than DGFont especially when given a small number of reference images. We observe that the variance of the two models declines with an increase in reference images number. 
It is because the contrastive representation model tends to learn robust and content-invariant style representation, which alleviates the influence of the variation of reference style images. 

\begin{table}[t]
\scalebox{0.95}{
\begin{tabular}{llllll}
\toprule
Methods  & L1 loss$\downarrow$& RMSE$\downarrow$& SSIM$\uparrow$& LPIPS$\downarrow$ & FID$\downarrow$\\
\midrule
DGFont & 0.0584&0.2166&0.7871&0.1204&41.69\\ 
DGFont+attn &0.0585&0.2170&0.7870&0.1190&39.67 \\
DGFont+attn+$\mathcal{L}_{sty}$ &\textbf{0.0567}&\textbf{0.2131}&\textbf{0.7910}&\textbf{0.1151}&\textbf{35.83} \\
\bottomrule
\end{tabular}}
\caption{ \textbf{Ablation study for DGFont++}. ``DGFont+attn+$\mathcal{L}_{sty}$" indicates the  model of DGFont++. }
\label{table:dgfont++}
\end{table}

\subsection{Analysis for FDSC Module}\label{sec:FDSC_analysis}
\subsubsection{FDSC vs skip-connection} 
We compare our proposed FDSC module with commonly used skip-connection \cite{DBLP:conf/miccai/RonnebergerFB15}. Skip-connection is often adopted to transfer feature maps with different resolutions directly from the encoder to the decoder, which is effective in semantic segmentation \cite{DBLP:conf/cvpr/JegouDVRB17,DBLP:conf/cvpr/LongSD15} whose content of inputs and outputs share the same structure. However, font generation requires a geometric deformation between content inputs and the corresponding generated images in the structure. To compare the FDSC module with skip-connection, We replace two FDSC modules with skip-connection in our proposed DGFont network. The comparison results are shown in Table \ref{unet}. We can observe that models with FDSC modules outperform models with skip-connection, which proves the effectiveness of FDSC.

\subsubsection{Visualization}\label{visualization}
\begin{figure}[t]
\begin{center}
   \includegraphics[width=\linewidth]{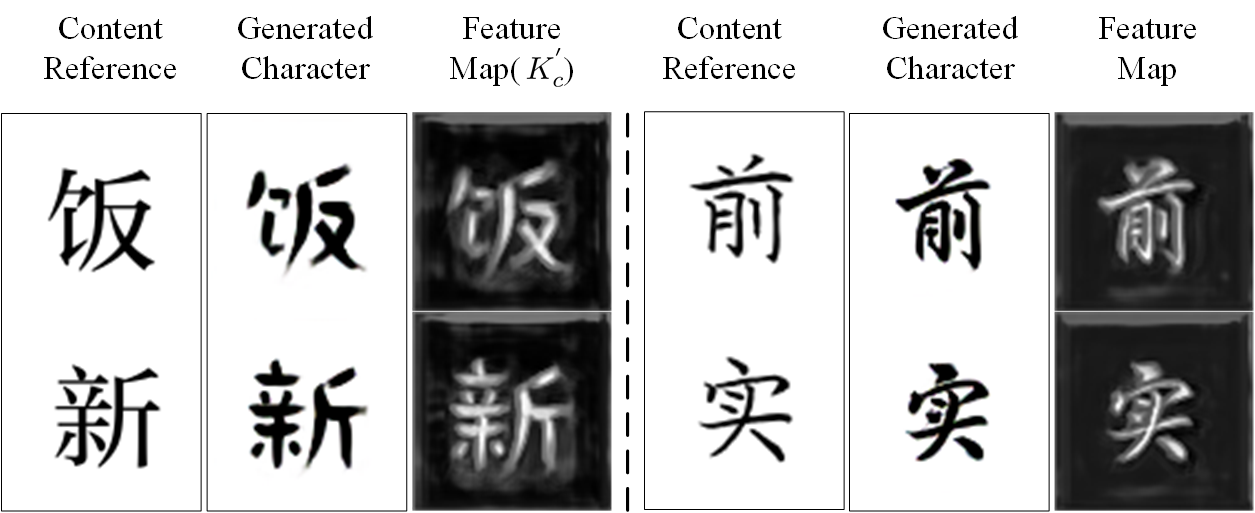}
\end{center}
   \caption{\textbf{Feature visualization}. We visualize the features $F^{'}_{c}$ generated from the FDSC module. For each case, from left to right: content reference characters, the corresponding generated characters, and the visualization of feature maps. For feature map images, the whiter area represents the larger activation value.}
\label{fig:5}
\end{figure}

In order to show the effectiveness of FDSC, we visualize the feature maps generated by the FDSC module. As shown in Figure \ref{fig:5}, the feature maps $F^{'}_{c}$ preserve the pattern of characters well, which helps generate a character with complete structure. On the other hand, we can observe that the FDSC module effectively transforms features extracted from the content encoder.

\begin{figure}[t]
\begin{center}
   \includegraphics[width=\linewidth]{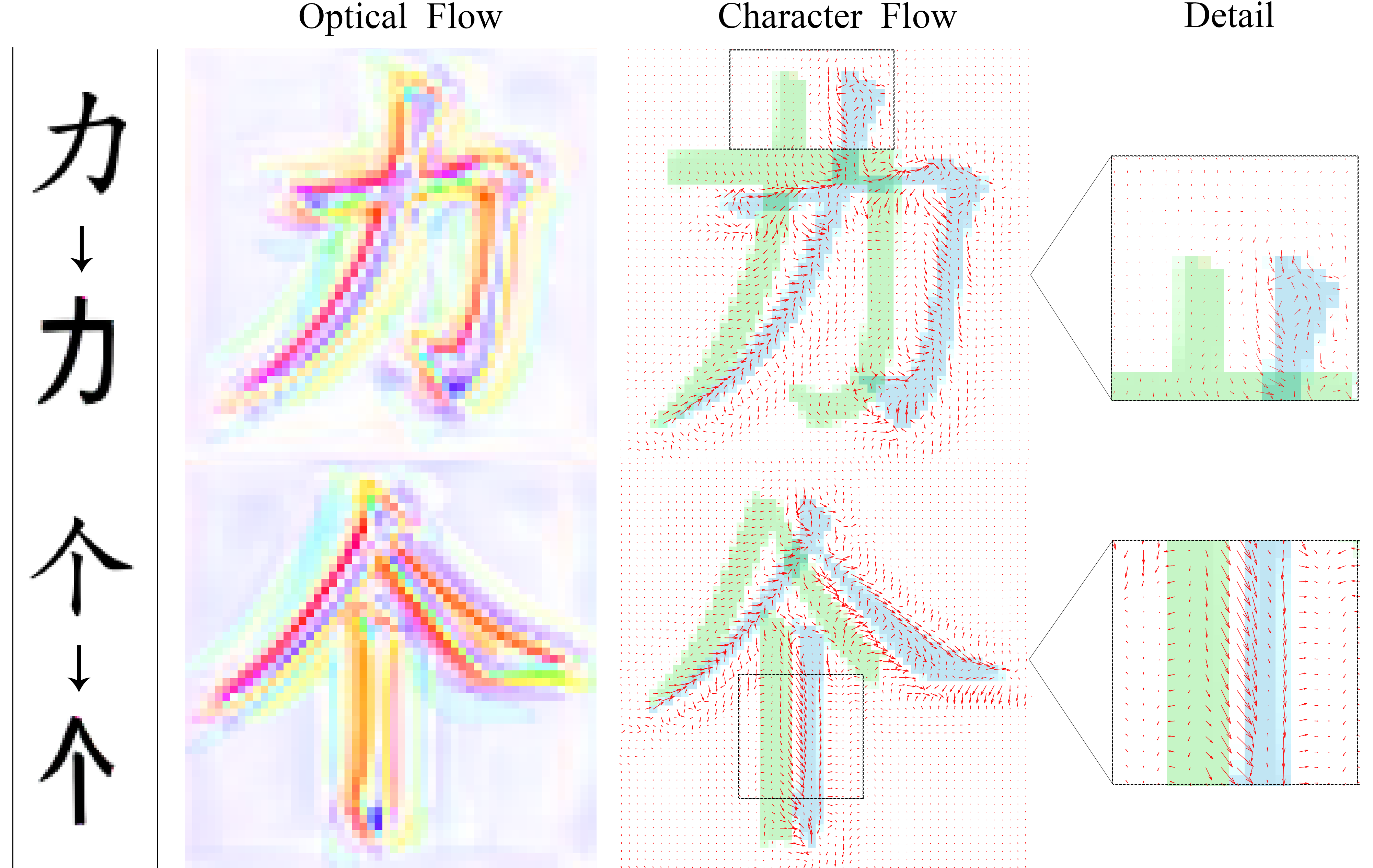}
\end{center}
   \caption{\textbf{The visualization of learned offsets}. First column: source images to generated images. Second column: the deformation flows of the estimated offsets $\Delta p$. Third column: visualization of the estimated offsets $\Delta p$ by quiver plot. Fourth column: zoomed-in details. The source and generated images are in blue and green respectively.}
\label{fig:6}
\end{figure}

In addition, we visualize the learned offsets from the FDSC module using optical flow and character flow respectively. To visualize the offsets, the kernel of deformable convolution in the FDSC module is set to 1$\times$1. 
As demonstrated in Figure \ref{fig:6}, we observe that the learned offsets mainly affect the character region. The offset value of the background tends to zero, which proves the usefulness of the proposed offset loss Eq. \ref{offsetloss}. In character flow, we can see that most of the offset vectors point from the stroke in target characters to the corresponding source stroke. The results show that in the convolution process, the sampling locations of target characters tend to shift to corresponding locations in the source character by the learned offsets.
\begin{table}[]
\begin{center}
\begin{tabular}{llllll}
\toprule
Method & L1 loss$\downarrow$& RMSE$\downarrow$& SSIM$\uparrow$& LPIPS$\downarrow$ & FID$\downarrow$ \\ \toprule
SC &    0.0641     &   0.2212   &   0.7252   &    0.1114   &  46.88   \\
FDSC                     &\textbf{0.0582}     &\textbf{0.2080}    & \textbf{0.7469}     &    \textbf{0.1006}   &  \textbf{46.39}   \\
\bottomrule
\end{tabular}
\end{center}
\caption{\textbf{Comparison with skip-connection (SC) proposed by U-Net} \cite{DBLP:conf/miccai/RonnebergerFB15} . We replace two FDSC modules with skip-connections and then compare the new model with the full model of DGFont.}
\label{unet}
\end{table}


\subsection{Style interpolation}
\begin{figure}[t]
\center
\includegraphics[width=\linewidth]{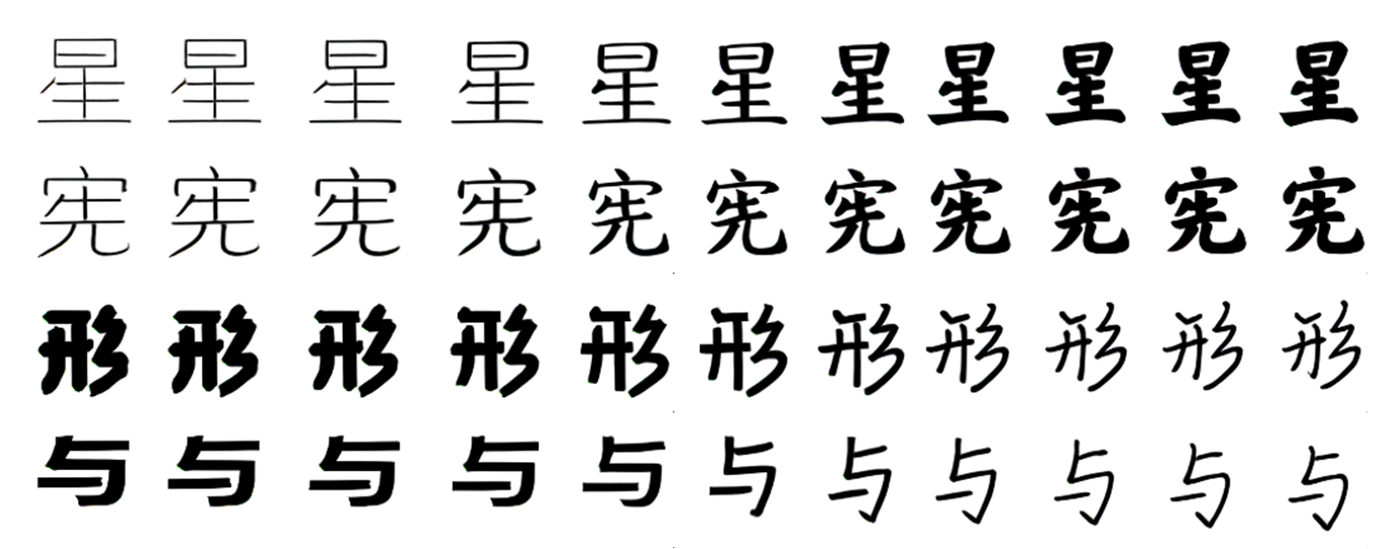}
\caption{ Style interpolation. Generated glyphs in each row correspond to identical content representations. The leftmost and rightmost images are generated given two fonts. Images in the middle are convex combinations of two fonts.}
\label{fig:style_inter}
\end{figure}

To demonstrate that our style representations are semantically meaningful, we provide the style interpolation results in Figure \ref{fig:style_inter}. We first extract style features from two different fonts, and linearly interpolate the style factors $Z_s$ to generate interpolated images. The interpolated factor is varied from 0 to 1 with an interval of 0.1. The results show that DGFont++ produces a smooth transition from one font to another and provides well-interpolated style features such as thickness, stroke, or joined-up writing while not hurting the character content.

\subsection{User Study}\label{user study}
\begin{figure}[t]
\begin{center}
   \includegraphics[width=1\linewidth]{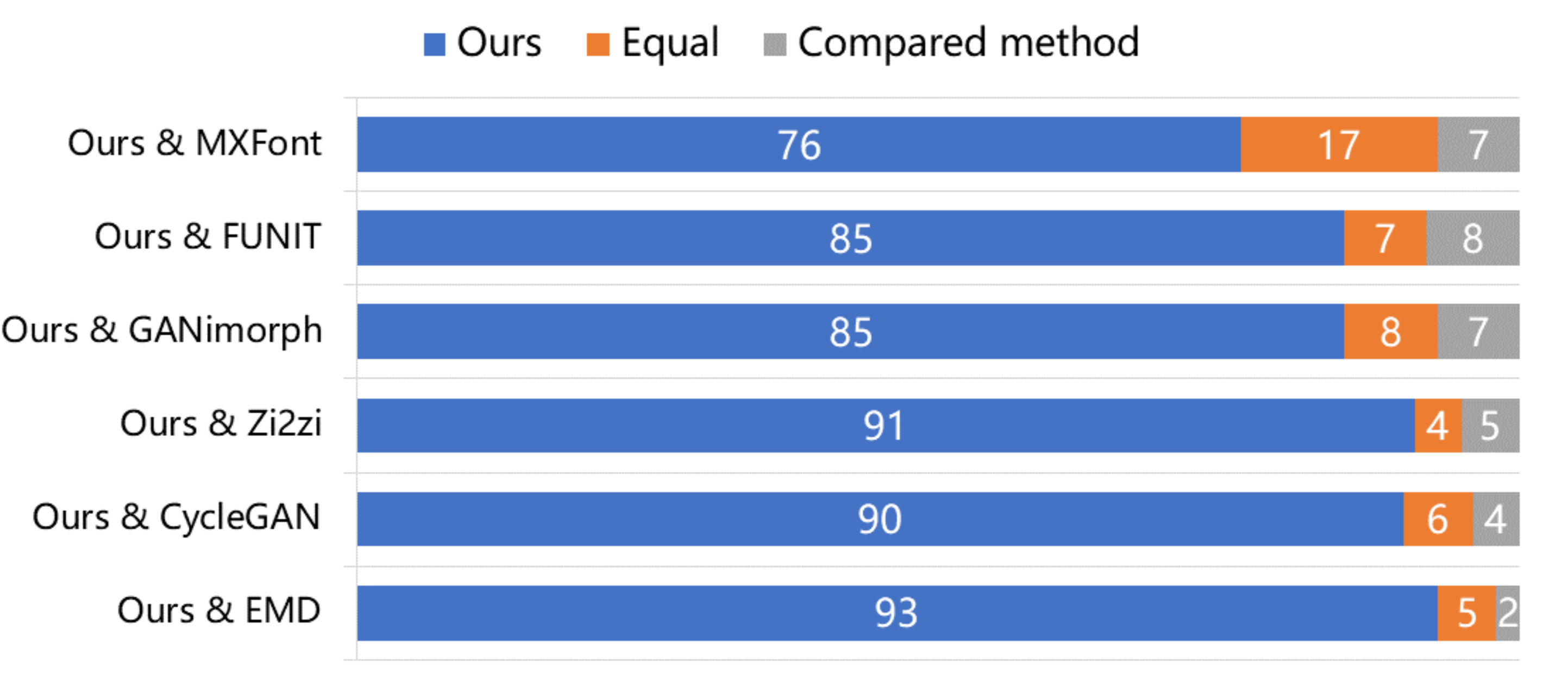}
\end{center}
   \caption{ \textbf{Results of user study}. For each compared method, we randomly sample 100 generated images for the participants' preferences test. The blue bar indicates the number of images that more participants prefer our results. The gray bar indicates the number of images that more participants prefer results from the compared methods. The orange bar indicates the number of images that get equal votes.}
\label{fig:user}
\end{figure}

To further compare the quality of images generated with other methods, we conduct an experiment on a human study by pairwise A/B tests. We compare our method with six compared methods.
For each comparison, we randomly select 100 characters from the test set to make 100 paired images generated by our method and another compared method. The participants are 10 people who use Chinese characters every day. The participants are asked to select a more similar image compared to ground-truth from each pair within seven seconds. For each pair, we choose the image with more votes as the judgment result. Figure \ref{fig:user} shows the participants' preference among the four tasks. We observe that more than 90 results of our methods outperform the results of Zi2zi, cycleGAN, EMD, which indicates that our method generates more realistic characters. Compared to FUNIT which gets impressive results in image-to-image translation tasks, our method still performs better on 85\% results. MXFont is the most recent method for font generation. 76\% results of our method are better than MXFont, which validates the superiority of our model over these state-of-the-art methods.
\section{Conclusion}
This paper proposes an effective unsupervised font generation model which is capable to generate realistic characters without paired images and can extend to unseen font well. We propose a Feature Deformation Skip Connection (FDSC) and FDSC-attn module to transfer the global and local deformable low-level spatial information to the mixer. Besides, we employ deformable convolution layers in the content encoder to learn style-invariant feature representations. Extensive experiments on font generation verify the effectiveness of our proposed model.

\appendices

\ifCLASSOPTIONcaptionsoff
\newpage
\fi

\bibliographystyle{IEEEtran}
\bibliography{egbib}
\end{document}